%% file: manuscript.tex
\def\year{2022}\relax
\documentclass[letterpaper]{article} % DO NOT CHANGE THIS
\usepackage{aaai22}  % DO NOT CHANGE THIS
\usepackage{times}  % DO NOT CHANGE THIS
\usepackage{helvet}  % DO NOT CHANGE THIS
\usepackage{courier}  % DO NOT CHANGE THIS
\usepackage[hyphens]{url}  % DO NOT CHANGE THIS
\usepackage{graphicx} % DO NOT CHANGE THIS
\usepackage{natbib}  % DO NOT CHANGE THIS AND DO NOT ADD ANY OPTIONS TO IT
\usepackage{caption} % DO NOT CHANGE THIS AND DO NOT ADD ANY OPTIONS TO IT
\DeclareCaptionStyle{ruled}{labelfont=normalfont,labelsep=colon,strut=off} % DO NOT CHANGE THIS
\usepackage{algorithm}
\usepackage{algorithmic}

\usepackage{newfloat}
\usepackage{listings}
\floatstyle{ruled}
\newfloat{listing}{tb}{lst}{}
\floatname{listing}{Listing}

\input{math_commands.tex}

\usepackage{booktabs}
\usepackage{makecell}

\usepackage{url}
\usepackage{xcolor}
\usepackage{amsmath}
\usepackage{amssymb}
\usepackage{multirow}
\usepackage{nccmath}

\newcommand{\ours}{Block-Skim}
\newcommand{\model}[2]{$\text{#1}_{\text{#2}}$}

\newcommand{\Fig}[1]{Fig.~\ref{#1}}

\newcommand{\Tbl}[1]{Tbl.~\ref{#1}}

\renewcommand{\paragraph}[1]{\vspace*{0.05cm}\noindent\textbf{#1}\hspace*{.1cm}}

\title{Block-Skim: Efficient Question Answering for Transformer}

\author {
    Yue Guan\textsuperscript{\rm 1,2},
    Zhengyi Li\textsuperscript{\rm 1,2},
    Zhouhan Lin\textsuperscript{\rm 1},
    Yuhao Zhu\textsuperscript{\rm 3},
    Jingwen Leng\textsuperscript{\rm 1,2},
    Minyi Guo\textsuperscript{\rm 1,2},
}
\affiliations {
    \textsuperscript{\rm 1} Shanghai Jiao Tong University\\ 
    \textsuperscript{\rm 2} Shanghai Qi Zhi Institute\\
    \textsuperscript{\rm 3} University of Rochester\\
    \{bonboru,hobbit,leng-jw\}@sjtu.edu.cn, 
    guo-my@cs.sjtu.edu.cn,
    lin.zhouhan@gmail.com, 
    yzhu@rochester.edu\\
}

\usepackage{bibentry}
\begin{document}
\maketitle

\input{sec/abstract}

\input{sec/introduction}
\input{sec/related_work}
\input{sec/motivation}

\input{sec/method}

\input{sec/experiment}

\input{sec/conclusion}

\appendix

\input{sec/appendix/attn_distrib}

\input{sec/acknowledgement}

\bibliography{bibliography}

\end{document}

%% file: math_commands.tex
\usepackage{amsmath,amsfonts,bm}

\def\eqref#1{equation~\ref{#1}}
\def\Eqref#1{Equation~\ref{#1}}
\def\1{\bm{1}}

\DeclareMathAlphabet{\mathsfit}{\encodingdefault}{\sfdefault}{m}{sl}
\SetMathAlphabet{\mathsfit}{bold}{\encodingdefault}{\sfdefault}{bx}{n}

%% file: sec/abstract.tex
\begin{abstract}

  Transformer models have achieved promising results on natural language processing (NLP) tasks including extractive question answering (QA).
  Common Transformer encoders used in NLP tasks process the hidden states of all input tokens in the context paragraph throughout all layers.
  However, different from other tasks such as sequence classification, answering the raised question does not necessarily need all the tokens in the context paragraph.
  Following this motivation, we propose \ours{}, which learns to skim unnecessary context in higher hidden layers to improve and accelerate the Transformer performance.
  The key idea of \ours{} is to identify the context that must be further processed and those that could be safely discarded early on during inference. 
  Critically, we find that such information could be sufficiently derived from the self-attention weights inside the Transformer model. 
  We further prune the hidden states corresponding to the unnecessary positions early in lower layers, achieving significant inference-time speedup.
  To our surprise, we observe that models pruned in this way outperform their full-size counterparts.
  % As a plugin to the Transformer-based QA models, \ours is compatible with other model compression methods without changing existing network architectures.
  \ours{} improves QA models' accuracy on different datasets and achieves $3\times$ speedup on \model{BERT}{base} model.

\end{abstract}

%% file: sec/introduction.tex
\section*{Introduction}
% \fixme{
%   trending of transformer architecture.\\
%   trending of larger model.\\
%   efficient transformers.\\
% }

%Compared to sequential models like LSTM~\citep{hochreiter1997long}, the MHA deals better with the long-term dependency through the attention connections, and is also easier to parallelize.
The Transformer model~\citep{vaswani2017attention} has pushed model performance on various NLP applications to a new stage by introducing multi-head attention (MHA) mechanism \citep{lin2017structured}. Further, the Transformer-based BERT~\citep{devlin2018bert} model advances its performances by introducing self-supervised pre-training and has reached state-of-the-art accuracy on many NLP tasks. This has made it at the core of many state-of-the-art models, especially in recent question answering (QA) models \citep{huang2020recent}.

%As a specific format of QA, extractive QA datasets follows the assumption that the answer is a continuous span of the provided document.
%On the contrast, natural language generation QA aims to produces a complete answer from scratch.

\begin{figure}
  \centering
  \includegraphics[width=\linewidth]{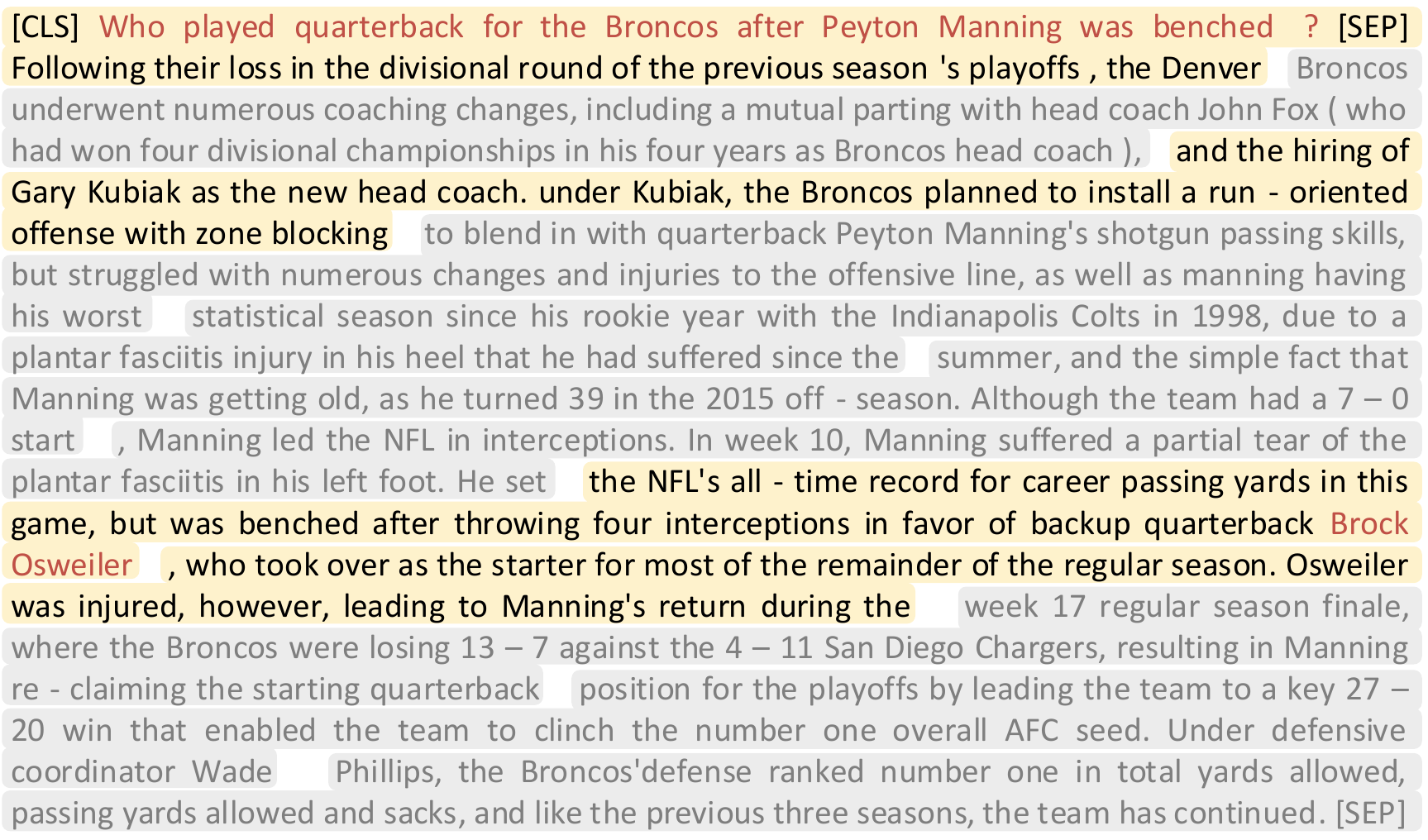}
  \caption{Example of \ours{} method on a query from the SQuAD dataset. The question and answer tokens are annotated in red. Only question and few evidence blocks are fully processed (annotated by yellow). And other blocks are skimmed for acceleration with the knowledge from attention weights (annotated by grey). Here the block size is 32 tokens.}
  \label{fig:posthoc}
\end{figure}

Our key insight for QA is that when human beings are answering a question with a passage as a context, they do \textit{not} spend the same level of comprehension for each of the sentences equally across the paragraph. Most of the contents are quickly skimmed over with little attention on it, which means that for a specific question most of the contents are \textbf{semantically redundant}. However, in the Transformer architecture, all tokens go through the same amount of computation, which suggests that we can take advantage of that by discarding many of the tokens early in the lower layers of the Transformer. This \emph{semantic level redundancy} sheds light on effectively reducing the sequence lengths at higher layers. Since the execution overhead of self-attention increases quadratically w.r.t. sequence length, this semantic level pruning could significantly reduce the computation time for long contexts.

To excavate the efficiency from this insight, we propose to first chop up the context into blocks, and then learn a classifier to terminate those less relevant ones early in lower layers by looking at the attention weights as shown in \Fig{fig:posthoc}. 
% By doing so, we reduce the overhead of processing irrelevant texts and accelerate the model inference.
% In the following sections, we show that giving more attention to the blocks containing answer improves the QA accuracy.
Moreover, with the supervision of ground truth answer positions, a model that jointly learns to discard context blocks as well as answering questions exhibits significantly better performance over its full-size counterpart.
Unfortunately, this also makes the proposed \ours{} method dedicated for extractive QA downstream task.
However, QA task is significant in real work production scenarios.
Moreover, our method lies in the trade-off space between generality, usability, and efficiency.
While sacrificing generality on applicable tasks, our proposed method is easy for adoption as it works as a plug-in for existing models.
Similarly, leveraging the QA-specific attention weight patterns makes \ours{} achieves better speedup results than other methods.

% feeding the attention mechanism with the knowledge of the answer position directly during training, the attention mechanism and QA model's accuracy are improved. 
% \fixme{this sentence is disconnected from the previous texts.}

In this paper, we provide the first empirical study on attention feature maps to show that an attention map could carry enough information to locate the answer scope.
We then propose \ours{}, a plug-and-play module to the transformer-based models, to accelerate transformer-based models on QA tasks. 
% \fixme{We need to highlight how we extract attention. Why is it non-trivial?} 
By handling the attention weight matrices as feature maps, the CNN-based \ours{} module extracts information from the attention mechanism to make a skim decision.
With the predicted block mask, \ours{} skips irrelevant context blocks, which do not enter subsequent layers' computation.
Besides, we devise a new training paradigm that jointly trains the \ours{} objective with the native QA objective, where extra optimization signals regarding the question position are given to the attention mechanism directly.

In our evaluation, we show \ours{} improves the QA accuracy and F1 score on all the datasets and models we evaluated.
Specifically, \model{BERT}{base} is accelerated for $3\times$ without any accuracy loss.

This paper contributes to the following 3 aspects.
\begin{itemize}
  \item We for the first time show that an attention map is effective for locating the answer position in the input.
  \item We propose \ours{}, which leverages the attention mechanism to improve and accelerate Transformer models on QA tasks. The key is to extract information from the attention mechanism during processing and intelligently predict what blocks to skim.
  \item We evaluate \ours{} on several Transformer-based model architectures and QA datasets and demonstrate its efficiency and generality.
\end{itemize}

% \fixme{we are the first to skim transformer}
% \fixme{we are the first to analyze the information exists in attention feature map}

%% file: sec/related_work.tex
 \section*{Related Work}

\paragraph{Recurrent Models with Skimming.}
The idea to skip or skim irrelevant sections or tokens of input sequence has been studied in NLP models, especially recurrent neural networks (RNN)~\citep{rumelhart1986learning} and long short-term memory network (LSTM)~\citep{hochreiter1997long}.
LSTM-Jump~\citep{yu2017learning} uses the policy-gradient reinforcement learning method to train an LSTM model that decides how many time steps to jump at each state.
They also use hyper-parameters to control the tokens before a jump, maximum tokens to jump, and the maximum number of jumping.
Skim-RNN~\citep{seo2018neural} dynamically decides the dimensionality and RNN model size to be used at the next time step.
In specific, they adopt two "big" and "small" RNN models and select the "small" one for skimming.
Structural-Jump-LSTM~\citep{hansen2018neural} uses two agents to decide whether to jump by a small step to the next token or structurally to the next punctuation.
Skip-RNN~\citep{DBLP:journals/corr/abs-1708-06834} learns to skip state updates and thus results in the reduced computation graph size.
The difference of \ours{} to these works is two-fold.
First, the previous works make the skimming decision based on the hidden states or embeddings during processing.
However, we are the first to analyze and utilize the attention mechanism for skimming.
Secondly, our work is based on the Transformer model~\citep{vaswani2017attention}, which has outperformed the recurrent type models on most NLP tasks.

\paragraph{Transformer with Input Reduction.}
% \zhouhan{Mention Universial Transformer.}
Unlike the sequential processing of the recurrent models, the Transformer model calculates all the input sequence tokens in parallel.
As such, skimming can be regarded as a reduction in sequence dimension.
Power-BERT~\citep{goyal2020power} extracts input sequence at a token level while processing.
During the fine-tuning process for downstream tasks, \citeauthor{goyal2020power} proposes a soft-extraction layer to train the model jointly.
Length-Adaptive Transformer~\citep{kim2020length} further extends Power-BERT by forwarding the rejected tokens to the final linear layer.
Funnel-Transformer~\citep{dai2020funnel} proposes a novel pyramid architecture with input sequence length dimension reduced gradually regardless of semantic clues.
For tasks requiring full sequence length output, such as masked language modeling and extractive question answering, Funnel-Transformer up-samples at the input dimension to recover.
DeFormer~\citep{cao2020deformer} propose to pre-process and cache the paragraphs at shallow layers and only concatenate with the question parts at deep layers.
Universal Transformer~\citep{dehghani2018universal} proposes a dynamic halting mechanism that determines the refinement steps for each token.
Different from these works, \ours{} utilizes attention information between question and token pairs and skims the input sequence at the block granularity accordingly.
Moreover, \ours{} does not modify the vanilla Transformer model, making it more applicable.

\paragraph{Efficient Transformer.}
There are also many efforts for designing efficient Transformers~\citep{zhou2020bert,wu2019lite,tay2020efficient}.
For example, researchers have applied well-studied compression methods to Transformers, such as pruning~\citep{guo2020accelerating}, quantization~\citep{wang2020q,guo2022squant}, distillation~\citep{sanh2019distilbert}, and weight sharing.
Other efforts focus on dedicated efficient attention mechanism considering its quadratic complexity of sequence length~\citep{kitaev2019reformer,beltagy2020longformer,zaheer2020big}.
\ours{} is orthogonal to these techniques on the input dimension reduction.
We demonstrate that \ours{} is compatible with efficient Transformers with experimental results.

%% file: sec/motivation.tex
\section*{Attention-based Block Relevance Prediction} \label{sec:motivation}

% \zhouhan{I kind of feel that most of the contents in section 3.1 can be deleted since they are basic knowledge. We'll just need Figure 2 and describing the attention patterns in the begining of 3.2}
\subsection*{Token-Level Relevance Analysis}

\textbf{Transformer.} 
The Transformer model adopts the multi-head self-attention mechanism and calculates hidden states for each position as an attention-based weighted sum of input hidden states.
The weight vector is calculated by parameterized linear projection query Q and key K as Equation~\ref{equ:attention}.
Given a sequence of input embeddings, the output contextual embedding is composed by the input sequence with different attention at each position,
\begin{equation} \label{equ:attention}
    \small
    Attention{(Q,K)} = Softmax({QK^T}/{\sqrt{d_k}}),
\end{equation}
where $Q, K$ are query and key matrix of input embeddings, $d_k$ is the length of a query or key vector.
As such, the attention weight feature map is often visualized as a heatmap demonstrating the information gathering relationship along tokens~\citep{kovaleva2019revealing}.
The model exploits multiple parallel groups of such attention weights, a.k.a. attention heads, for attending to information at different positions.

\begin{figure}[t]
  \centering
  \includegraphics[width=\linewidth]{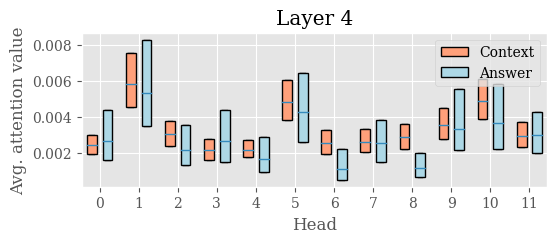}
  \includegraphics[width=\linewidth]{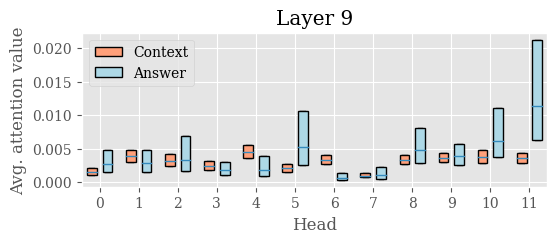}
  \caption{Attention weight value distribution comparison on the answer and irrelevant tokens. The attention heatmaps are profiled on the development set of SQuAD dataset with a \model{BERT}{base} model with 12 layers and 12 attention heads per layer. The full results are shown in appendix.}
  \label{fig:avg_attn}
    \vspace*{-0.2cm}
\end{figure}

Extractive QA is one of the ultimate downstream tasks in the NLP.
Given a text document and a question about the context, the answer is a contiguous span of the text.
To predict the start and end position of the input context given a question, the embedding of each certain token is processed for all the layers in the Transformer encoder model.
% On top of the Transformer encoder, each embedding is given to a linear classification model to compute the score of such position.
% This means all tokens are processed for all the transformers layers.
% However, most sentences and contents are not relevant to the question and do not contribute to comprehend the context.
% Under the characteristic of extractive QA problem that answer spans are contiguous, it is possible for us to narrow the answer position by block granularity.
% As such, we manage to figure out the important part for fully processing and skim the others.
In many end-to-end open-domain QA systems, information retrieval is the preceding step at the coarse-grained passage or paragraph level for filtering out irrelevant passages.
With the characteristic of the extractive QA problem where answer spans are part of the passage, our question is that whether we can apply a similar filtering technique at fine-grained granularity during the Transformer model inference.

% In other words, are the attention weights effective for distinguishing the answer blocks?
% In other word, does attention head attend differently over answer sentences.

In this work, we propose to augment the attention mechanism with the ability to predict the relevance of contextual tokens without modifying the original Transformer model.
Prior work~\cite{goyal2020power} shows that attention strength is a good indicator for answer tokens.
However, we analyze the attention weight distribution of a trained \model{BERT}{base} model trained with SQuAD~\citep{rajpurkar2016squad} dataset and find that the attention weights of multi-head attention only have noticeably patterns at the late layers. 

\Fig{fig:avg_attn} compares the attention weights at Layer 4 and 9 in the trained \model{BERT}{base} model.
The tokens are classified to answer tokens or irrelevant tokens with the labels from the dataset.
At late layers like Layer 9, the attention weights of answer tokens are significantly larger than those of irrelevant tokens.
However, at early layers like Layer 4, the attention weight strength is indistinguishable for answer tokens and irrelevant tokens.
For a better latency reduction, it is desirable to find irrelevant tokens as early as possible. 
However, using the attention weight value as the relevance criterion could be problematic at early layers.
 
%The results in \Fig{fig:avg_attn} show that different attention heads have different functionality.
%For example, No.0, No.10 or No.11 attention heads at layer 9 place much higher attention value at answer tokens, while No.6, No.8 are the opposite.
%% In other word, this specific head acquires the exact answer tokens.
%% These observation still holds for bottom layers though being less significant, for instance, heads No.7 and No.9 at 4th layer.
%This demonstrates that attention value has distinct distribution on answer tokens.
%So it is possible to locate the answer area with the attention weight heatmap.
%Because the distribution of the attention values is non-trivial.

\begin{figure}[t]
  \centering
  \includegraphics[width=\linewidth]{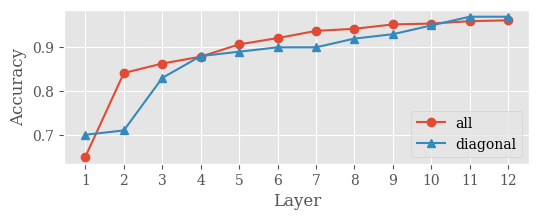}
  \caption{Accuracy of CNN model predicting whether a block contains answer with attention weight as input. The CNN is feed with either an all attention weight heatmap or only the diagonal block region.}
  \label{fig:sentence_retrieval}
  \vspace*{-0.2cm}
\end{figure}

\subsection*{CNN Based Block Relevance Prediction}

Given the complex pattern of attention weights, we propose to use a CNN-based feature extractor to process the attention heatmaps as input image channels and predict the relevance of each token.
To amortize the processing overhead, we split the input sequence $X=(x_0,x_1,...,x_i)$ into $i/k$ exclusive blocks $block_j=(x_{j\times k},x_{j\times k+1}...,x_{j \times k+(k-1)})$, where $k$ is the block size, i.e. tokens included in the continuous input span.
The relevance of a block is defined as whether it contains the exact final answer.
As such, our goal is to figure out the blocks' relevance and skim the irrelevant ones during Transformer inference.

% To filter the input sequence at fine-grained granularity, we split the input sequence to fixed-length blocks to predict whether each block contains the answer.
% Then the attention heatmaps from a Transformer model is feed to a CNN model with each attention head representing an input channel.

% We propose the \ours{} to accelerate the question answering task without degrading the answer accuracy.
% Unlike the Transformer-based model that uses all input tokens throughout all the layers, \ours{} accurately identifies the irrelevant contexts for the question in the early layers, and remove those irrelevant contexts in the following layers.
% As such, our model reduces the computation requirement and enables fast question answering.
% In \Sec{sec:background}, we have shown the feasibility of exploiting the attention relationship among tokens to identify tokens that do not contain answers to the question.
% However, this naive approach significantly hurts the QA accuracy as we will show later.
% As such, we propose an end-to-end learnable feature extractor that can be jointly trained with the original QA task for the better accuracy.

\begin{figure*}[t]
  \centering
  \includegraphics[width=\linewidth]{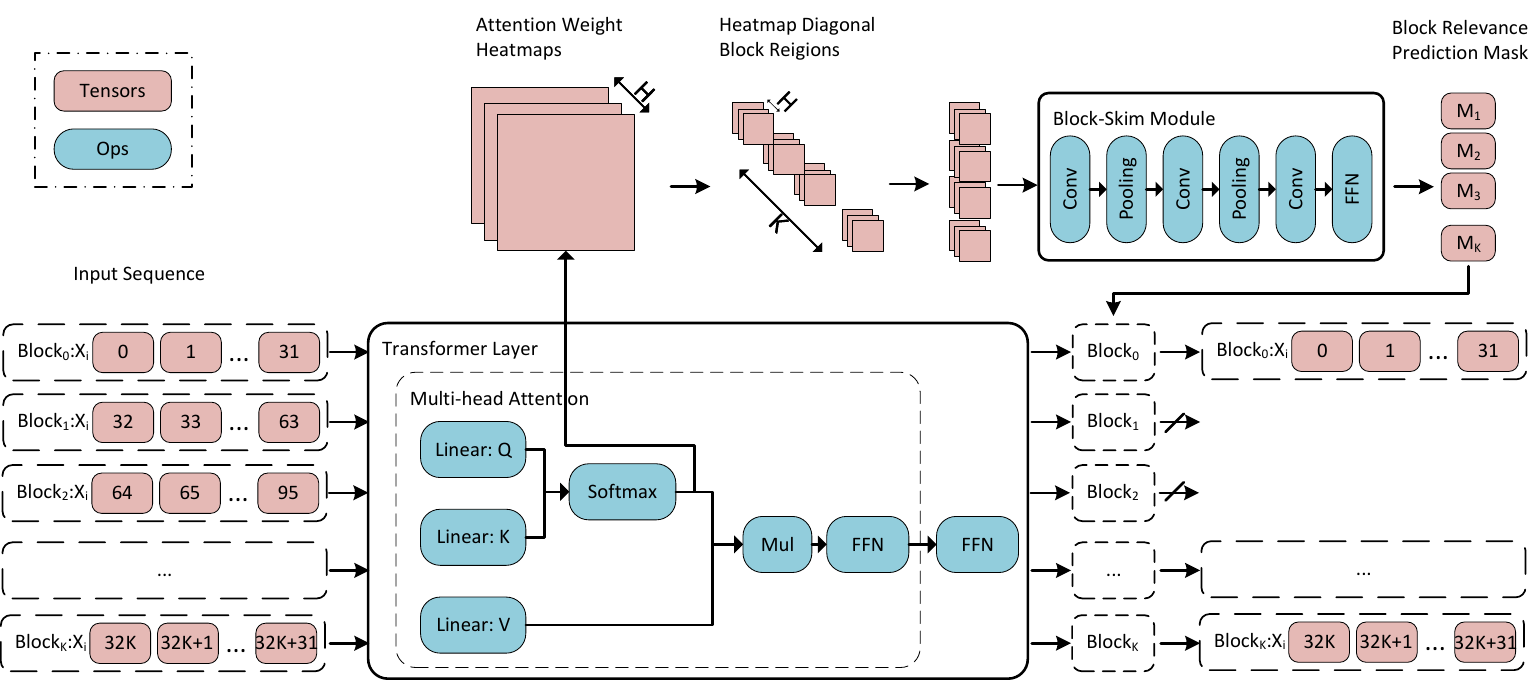}
  \caption{The overall schematic of \ours{} and the architecture of the CNN model. Here we take block size 32 as example. The total number of blocks and attention heads are K and H. We only show the main operations for simplicity.}
  \label{fig:cnn_arch}
\end{figure*}

\Fig{fig:cnn_arch} shows the details of how we extract the attention information from the Transformer and feed them into the CNN model.
In the CNN module, we use two $3\times3$ convolution and one $1\times1$ convolution, all of which use the \texttt{ReLU} operation~\citep{hahnloser2001permitted} as the activation function.
We insert a $2\times2$ average pooling layer for the first two $3\times3$ convolutional layers to reduce the feature map size.
In addition, we also use two batch normalization layers~\citep{ioffe2015batch} to improve the prediction accuracy.
To locate the answer context blocks, we use a linear classification layer to calculate the score for each block.
The module outputs a block-level prediction mask that corresponds to the relevance of a block of input tokens to the question. 

%With the insight described in \ref{sec:motivation}, we propose a plugin rejection architecture to the transformer layer to terminate the processing of irrelevant input blocks earlier.
%Though handcrafted feature achieves descent locating accuracy, we propose a parameterized feature extractor which better captures the attending information.
%To utilize the locality of attention of adjacent positions, we regard the attention weight matrix of a self-attention head as a feature map and convolve it to extract high-level information.

% Formally, we denote the input sequence of a Transformer layer as $X=(x_0,x_1,\dots,x_n)$, and the attention matrices of this layer as $Attention(X)$.
% Given the attention output of a Transformer layer, the $k_{th}$ block prediction result $M$ is represented as $M = BlockSkim (Attention(X))$, where the main functions of \ours{} is expressed as \Fig{fig:arch}(c). 

This model is trained with all attention heatmap profiled from the same set of heatmap data as described before. 
The prediction accuracy is shown as \Fig{fig:sentence_retrieval}.
In general, the model achieves decent accuracy demonstrating that a CNN model is capable to extract the attending behavior information and locate the answer.
Intuitively, the CNN models with higher layer attention heatmaps have better performance.
It suggests that the backbone model becomes more convinced on question answering when it gets deeper.

\subsection*{Simplifying CNN Predictor with Diagonal Attention}
\label{subsec:diagonal_cnn}

% Further, we reduce the input attention heatmap region to the diagonal block, which represents the internal attention within the block.
% Further, we reduce the input attention heatmap region to the diagonal block to reduce the computational complexity.

The above method of feeding the whole attention feature map to the CNN predictor has a  major problem, which is the predictor needs to deal with the variable size of the attention feature map.
As such, we simplify the input to the CNN model with only attention from its diagonal region.
In specific, we only feed the diagonal heat-map region as the input representation for each input sequence block, as expressed in~\Fig{fig:cnn_arch}.

Our \textbf{hypothesis} is that the diagonal region of the attention heat map contains sufficient information to identify the block relevance.
% This is motivated by an important characteristic of the attention mechanism which has been discussed by many previous works.
Because previous works~\citep{clark2019does, guan2020far} show that the attention mechanism has several fixed patterns, that is, diagonal, stride, block, or dense types.
And all of these patterns can be easily recognized with only the diagonal region.
% Based on these findings, there are many research works aiming to reduce or approximate the calculation of vanilla attention mechanism.

% Based on their findings, we take the diagonal parts as a good representation for each input sequence block.
Similarly, we optimize CNN models with reduced heatmap and the result is shown as Fig.~\ref{fig:sentence_retrieval}. 
As we can see, the models achieve similar prediction accuracy compared with using a whole attention weight heatmap.
The result justifies our hypothesis that it is possible to use the diagonal information from attention heatmaps to predict the answer relevance.
By doing so, the computation complexity is also reduced dramatically as the input size is much smaller.

The above finding confirms our hypothesis that the diagonal attention weight indeed carries information for figuring out answer positions.
This motivates us to utilize such attention information to narrow the possible answer position along with the processing of the input sequence.
In the next section, we introduce our design that uses a plug-and-play end-to-end learning module to extract useful information from the attention weights for skimming decisions.

%% file: sec/method.tex
\section*{Transformer with \ours{}} \label{sec:methods}
% \fixme{
%   feature extractor archtecture\\
%   joint training method\\
%   rejection and forward setting
% }

% \fixme{re-write this section with clear math notations}

The previous section shows the feasibility of using the attention weights to predict the relevance of token blocks.
However, naively using the predictor can lead to significant degradation of the QA task accuracy.
Because the block relevance predictor is only trained with the answer labels, it could fail in the multi-hop QA task, which requires information beyond the answer labels.
To solve this problem, we propose an end-to-end multi-objective joint training paradigm.
Then during inference time, the prediction of the \ours{} model is augmented to filter the input sequence for acceleration.
This causes a mismatch between training and inference models.
However, skimming blocks during training makes joint training unstable.
And our experimental results demonstrate that this mismatch is negligible.
We give a detailed demonstration of the proposed joint training paradigm and inference process as follows.

%we propose \ours{} to identify the irrelevant parts of the input paragraph during Transformer processing time.
% \ours{} extracts the attention weights during Transformer inference and feed the diagonal parts to a CNN model to predict answer appearance.
%To better recognize the necessary evidences for the answer, especially in a multi-hop scenario, the appended \ours predictor is end-to-end optimized with the backbone model without any loss of performance.

\subsection*{Single-Task Multi-Objective Joint Training} \label{sec:method:objective}

Following the previous experiments, we append the aforementioned CNN models to each layer to predict the blocks' relevance and optimize them together with the backbone Transformer model.
As such, there are two types of classifiers in the model augmented with \ours{} module.
The first is the original QA classifier at the last layer and the second is the block-level relevance classifier at each layer.
These two classifiers optimize the same downstream task of predicting the answer position with an identical target label. However, they are fed with a different type of loss objectives, that is, the QA objective with Transformer output embeddings and the Block-Skim objective with attention weights. We jointly train these classifiers so that the training objective is to minimize the sum of all classifiers' losses.

The loss function of each block-level classifier is calculated as the cross-entropy loss against the ground truth label whether a block contains answer tokens or not.
Equation~\ref{eq:label} gives the formal definition.
The total loss of the block-level classifier $\mathcal{L}_{BlockSkim}$ is the sum of all blocks that only contain passage tokens.
The reason is that we only want to throw away blocks with irrelevant passage tokens instead of questions.
Blocks that have question tokens are not used in the training process.
\begin{equation} \label{eq:label}
  \begin{aligned}
    \mathcal{L}_{BlockSkim}=&\sum_{m_i \in \{\texttt{passage blocks}\}}{CELoss(m_i,y_i)}\\
    y_i = &
    \begin{cases}
      1 & \text{, block i has answer tokens}\\
      0 & \text{, block i has no answer tokens}\\
    \end{cases}    
  \end{aligned}
\end{equation}

For the calculation of the final total loss $\mathcal{L}_{total}$, we introduce two hyper-parameters in \Eqref{eq:totalloss}.
We first use a harmony coefficient $\alpha$ so that different models and settings could adjust the ratio between the QA loss and block-level relevance classifier loss.
It is decided by grid search on the development set.
We then use the balance factor $\beta$ to adjust the loss from positive and negative relevance blocks because there are typically many more blocks that contain no answer tokens (i.e., negative bocks) than the blocks that do contain answer tokens (i.e., positive bocks).
This hyper-parameter selection will be explained in detail in experiments setup.
\begin{equation} \label{eq:totalloss}
  \mathcal{L}_{total} = \mathcal{L}_{QA} + \alpha  \sum_{ i_{th}\text{ layer}}{(\beta \mathcal{L}_{BlockSkim}^{i,y=1}+ \mathcal{L}_{BlockSkim}^{i,y=0})}
\end{equation}
Our \ours{} is a convenient plugin module owing to the following two reasons.
First, it does not affect the backbone model calculation, because it only regularizes the attention value distribution with extra parameters to the backbone model.
In other words, a model trained with \ours{} can be used with it removed.
Second, the introduced \ours{} objective neither needs an extra training signal nor reduces the QA accuracy.
In fact, we will show that the extra gradient signal feeding to the attention improves the original QA accuracy.
% Meanwhile, we will demonstrate that \ours{} works well with Transformer-based Roberta~\citep{liu2019roberta}, which has a different pre-training objective and sequence encoding, and ALBERT~\citep{lan2019albert}, which shares weights among layers for reducing the model size.

% \zhouhan{I think we won't need a whole paragraph here arguing for multi-hop QA. We just need to mention briefly here and elaborate in the experiment section.}

\paragraph{Multi-hop QA.}
Our joint training approach can also address the challenge in the multi-hop QA tasks~\citep{yang2018hotpotqa}, where deriving answers requires multiple pieces of evidence and reasoning.
Although the block relevance prediction only uses the answer label signal, the original QA task ensures that evidence needs to be kept.
In other words, the evidence reasoning information is encoded implicitly in the contextual embeddings.
To illustrate such a point, we perform an ablation study that incorporates the evidence label in the \ours{} predictor training.
The predictor accuracy does not improve with the additional evidence label, which confirms the effectiveness of our single-task multi-objective joint training.

\subsection*{Inference with Block-Skim}

We now describe how to use the \ours{} to accelerate the QA task inference.
Although we add the block-level relevance classification loss in the joint training process, we do not actually throw away any blocks because it can skip answer blocks and the QA task training becomes unstable.
% As such, the block-level relevance classification loss can be viewed as a regularization method for the QA training as we force attention heads to better distinguish the answer blocks and non-answer blocks.
% Because the \ours{} module works as an add-on component to the vanilla Transformer and does not change its computation, a jointly trained model can work just normally by removing the \ours{} module, or work in the skimming mode with decisions from the module.
% As we show later, the proposed skim objective is compatible with vanilla QA objective without any loss of performance.
% It actually improves the accuracy of the original QA task.
% In the above joint training process, we insert a \ours{} module in every layer without actual skimming.
However, we only augment block reduction with the \ours{} module during the inference for saving computation and avoiding heavy changes to the underlying Transformer. 
% As such, the layers to augment is a design choice in our model.
% Here we simply trigger it at each layer and leave the searching policy as future work.
%As such, the computation of the model can be regarded as pruned at input dimension.
During inference computation, we split the input sequence by the block granularity, which is a hyper-parameter in our model.
The model skips a set of blocks according to the skimming module results for the following layers.
% We would like to emphasize that \ours{} training process does not discard any blocks because if a block with answer tokens is discarded, the original QA training becomes unstable.
%blocks are not rejected during the fine-tuning optimization process.
%We only train the \ours{} plugin module parameters.
With those design features, \ours{} works as an add-on component to the original Transformer model and is compatible with many Transformer variant models as well as model compression methods.

% To maintain compatibility with the original Transformer model, we forward the skipped blocks directly to the last layer for the QA classifier.

%As such, the position of the BSM-augmented layer and the size of the block belong to hyper-parameters in our model.
%Besides computation saving, augmenting only a single layer with the BSM module also leads to minimal changes for the original model.
%Although input sequences can be pruned right after the multi-head self-attention operation, we insert this operation between transformers layers.
%This makes the architecture of the transformer unchanged and \ours{} works as a plugin module.
%On the other hand, the rejected blocks are bypassed directly to the final QA discriminator.
%With the retrieval classification result, it is possible for the model to narrow the position of the final answer and reject irrelevant blocks from the successive computation.

% \fixme{move this paragraph to appendix}
We provide an analytical model to demonstrate the latency speedup potential of \ours{}.
Suppose that we insert the \ours{} module to a vanilla model with the total $L$ layers, and a portion of $m_i$ blocks remains for the following layers after layer $i$.
The ideal processing complexity of one token for one Transformer layer is noted as $T_{layer}$.
Here we make an approximation that the computation complexity is linear to the sequence length $N$.
This is a conservative approximation because the attention mechanism is $O(N^2)$.
The performance speedup is formulated by Equation~\ref{eq:speedup} if we ignore the computation overhead of \ours{}.
%If we don't count the overhead of block retrieval module, the computation speedup is 
In fact, the computation of a single \ours{} module is smaller than Transformer layers for 100 times.
For example, when $\sum_{m_k \in \{\texttt{passage blocks}\}}m_k=0.9$, it means $10\%$ of tokens are skimmed each layer.
This skimming decision will result in a total speedup ratio of $1.86\times$.
% \begin{equation} \label{eq:speedup}
%   speedup = \frac{L \cdot N \cdot Layer}{l \cdot N \cdot Layer+(L-l) \cdot N \cdot k \cdot Layer} = \frac{L}{l+(1-k)L}
% \end{equation}

\begin{equation} \label{eq:speedup}
\begin{medsize}
  \begin{aligned}[b]
  Speedup & = \frac{T_{\tiny{Vanilla}}}{T_{\tiny{\ours{}}}} \\
          \scriptsize & =  \frac{L\cdot N \cdot T_{layer}}{\sum_{i=0}^{L}(\prod_{j=0}^{i}\sum_{m_{j,k} \in \{layer_j\}}m_{j,k}\cdot N) \cdot T_{layer}}  \\
          & = \frac{L}{\sum_{i=0}^{L}\prod_{j=0}^{i}\sum_{m_{j,k} \in \{layer_j\}}m_{j,k}} \
  \end{aligned}
\end{medsize}
\end{equation}

%% file: sec/experiment.tex
\begin{table*}[]
\center
\huge
\resizebox{\linewidth}{!}{
\begin{tabular}{c|cc|cc|cc|cc|cc|cc|cc}
\toprule
\multirow{2}{*}{Datasets}    & \multicolumn{2}{c|}{SQuAD} & \multicolumn{2}{c|}{HotpotQA} & \multicolumn{2}{c}{NewsQA} & \multicolumn{2}{c|}{NaturalQuestions} & \multicolumn{2}{c}{TriviaQA} & \multicolumn{2}{c|}{SearchQA} & \multicolumn{2}{c}{Avg.} \\ \cmidrule(lr){2-15} 
                            & F1          & Speedup      & F1           & Speedup        & F1          & Speedup      & F1               & Speedup            & F1           & Speedup       & F1           & Speedup        & F1         & Speedup      \\ \cmidrule(lr){1-15} 
Balance Factor              & \multicolumn{2}{c|}{20}    & \multicolumn{2}{c|}{20}       & \multicolumn{2}{c}{30}     & \multicolumn{2}{c|}{30}               & \multicolumn{2}{c}{100}      & \multicolumn{2}{c|}{150}      & \multicolumn{2}{c}{-}    \\ \cmidrule(lr){1-15} 
Vanilla BERT                & 88.32       &    $1\times$   & 74.39        &    $1\times$    & 66.57       &    $1\times$     & 78.85            &      $1\times$       & 72.61        &   $1\times$          & 79.93        &     $1\times$    & 76.78      &    $1\times$     \\
Block-Skim Training         & 88.92       &    $1\times$      & 74.88        &      $1\times$     & 67.76       &     $1\times$      & 78.98            &      $1\times$        & 73.29        &        $1\times$   & 80.32        &      $1\times$      & 77.36      &      $1\times$      \\ \cmidrule(lr){1-15} 
Block-Skim Inference        & 88.52       & $3.01\times$         & 74.47        & $2.28\times$           & 65.14       & $2.53\times$         & 78.48            & $2.56\times$               & 72.80        & $1.81\times$          & 79.84        & $3.17\times$           & 76.54      & $2.56\times$         \\
Deformer                    & 87.2        & $3.1\times$          &  -            &       -         &         -    &        -      &  -                &            -        &     -         &    -           &        -      &         -       &      -      &   -           \\
Length-Adaptive Transformer & 88.7        & $2.22\times$         &           -   &       -         &      -       &  -            &  -                &        -            &       -       &     -          &          -    &      -          &        -    &   -           \\ \bottomrule
\end{tabular}
}
\caption{Validation F1 score and FLOPs speedup of \model{BERT}{base} model evaluated on different QA datasets. The balance factor is determined by calculating the block number distribution on training set. }
\label{table:speedup_datasets}
\end{table*}
\section*{Evaluation} \label{sec:experiment}

\subsection*{Experimental Setup} \label{sec:experiment:setup}

% We evaluate the model accuracy improvement with rejection loss introduced on different pre-trained models including bert encoder, Roberta and Albert.
% During the fine-tuning process, the hyper parameters, e.g. learning rate, batch size, warm-up proportion are selected following the origin papers.
% To evaluate the speedup of proposed input rejection method, we measure the average processing time on the whole evaluation set.
% The speedup is evaluated on CPU and RTX1080 GPU.
% Because when a block containing the correct answer is rejected unexpectedly, the model can never give the right answer.
% The reject classifier threshold can be tuned as a trade-off between rejection speedup and accuracy.

\paragraph{Dataset.} 
We evaluate our method on 6 extractive QA datasets, including SQuAD 1.1~\citep{rajpurkar2016squad}, Natural Questions~\citep{kwiatkowski2019natural}, TriviaQA~\citep{joshi2017triviaqa}, NewsQA~\citep{trischler2016newsqa}, SearchQA\citep{dunn2017searchqa} and HotpotQA~\citep{yang2018hotpotqa}.
% All these datasets varies in context length, document source, and evidence independence.
HotpotQA provides questions that require multi-hop reasoning to answer with supporting facts.
The diversity of these datasets such as various passage lengths and different document sources lets us evaluate the general applicability of the proposed method.
% Following the BERT~\citep{} setting for QA, we use the transformer encoder to process all these datasets.

\paragraph{Model.} 
We follow the setting of the BERT model to use the structure of the Transformer encoder and a linear classification layer for all the datasets.
As previously explained, \ours{} works as an add-on module to the vanilla Transformer, and therefore is generally applicable to all Transformer-based models, as well as model compression methods.
To illustrate this point, we apply the \ours{} method to two BERT models with different size settings.
We evaluate the base setting with 12 heads and 12 layers, as well as the large setting with 24 layers and 16 heads as described in prior work~\citep{devlin2018bert}.

\paragraph{Model Compression Methods.} We conduct the following model compression methods on \model{BERT}{base} models to demonstrate the compatibility of our \ours{}.
\begin{itemize}
    \item \textbf{Distillation.} Knowledge distillation uses a teacher model to transfer the knowledge to a smaller student model. Here we adopt DistilBERT~\citep{sanh2019distilbert} setting to distill a 6-layer model from the \model{BERT}{base} model.
    \item \textbf{Weight Sharing.} By sharing weight parameters among layers, the amount of weight parameters reduces. Note that weight sharing does not impact the computation FLOPs (floating-point operations). We evaluate \ours{} on ALBERT~\citep{lan2019albert} that shares weight parameters among all layers. 
    \item \textbf{Pruning.} Instead of conventional weight pruning techniques, we evaluate head pruning~\citep{michel2019sixteen} that is specific to the attention mechanism in Transformer models.
    The pruning of attention heads reduces the input feature size to the \ours{} module.
    We prune $50\%$ of attention heads based on the attention head importance criterion introduced in prior work~\citep{michel2019sixteen}.
\end{itemize}

\paragraph{Input Dimension Reduction Baselines.} We also compare with input dimension reduction methods Deformer and Length-Adaptive Transformer. Deformer\citep{cao2020deformer} pre-process and caches the context paragraphs at early layers to reduce the actual inference sequence length. Length-Adaptive Transformer~\cite{kim2020length} is a successive version of Power-BERT which forwards the tokens rejected to the final layer by attention strength.

\paragraph{Training Setting.} 
We implement the proposed method based on open-sourced library from \citet{Wolf2019HuggingFacesTS}\footnote{The source code is available at \url{https://github.com/ChandlerGuan/blockskim}.}.
For each baseline model, we use the released pre-trained checkpoints~\footnote{We use pre-trained language model checkpoints released from \url{https://huggingface.co/models}.}.
We follow the training setting used by \citet{devlin2018bert} and \citet{liu2019roberta} to perform the fine-tuning on the above extractive QA datasets.
We initialize the learning rate to $3e-5$ for BERT models and $5e-5$ for ALBERT with a linear learning rate scheduler.
For SQuAD dataset, we apply batch size 16 and maximum sequence length 384.
And for the other datasets, we apply batch size 32 and maximum sequence length 512.
We perform all the experiments reported with random seed 42.
We train a baseline model and \ours{} model with the same setting for two epochs and report accuracies from MRQA task benchmark for comparison.
We use four V100 GPUs with 32~GB memory for the training experiments. 
%The training is performed on 2 or 4 32GB V100 GPUs differently without any influence on the result.

The balance factor $\beta$ is determined by block sample numbers and reported in \Tbl{table:speedup_datasets}.
The harmony factor $\alpha$ is $0.01$ for ALBERT and $0.1$ for all the other models we used.
It is determined by hyper-parameter grid search from $1e-3$ to $10$ with a step of $\times 10$.
% We set the hyper-parameter $\beta$ to 4 for all experiments and $\alpha$ to 1 except ALBERT.
% We use the $\alpha$ value of 0.05 for ALBERT.
% In the ALBERT model, the parameters of transformer layers are shared but \ours{} modules in our method do not share parameters.
% As such, we decrease the loss from \ours{} to prevent model over-fitting and its impact on the QA task parameters.

We use the inference FLOPs as a general measurement of the model computational complexity on all platforms.
We use TorchProfile\citep{torchprofile} to calculate the FLOPs for each model and normalize the results as a ratio to \model{BERT}{base}.
\begin{table}[]
\huge
\resizebox{\linewidth}{!}{
\begin{tabular}{ccccccccc}
\toprule
Seed                                 &    & 0     & 1     & 2     & 3     & 4     & Avg.  & Std. \\ \cmidrule(lr){1-9}
\multirow{2}{*}{Vanilla} & EM & 80.95 & 81.08 & 80.98 & 81.06 & 80.96 & 81.00 & 0.06 \\
                                     & F1 & 88.32 & 88.44 & 88.59 & 88.44 & 88.42 & 88.44 & 0.10 \\ \cmidrule(lr){1-9}
\multirow{2}{*}{Block-Skim}      & EM & 81.52 & 81.25 & 81.24 & 81.51 & 81.84 & 81.47 & 0.25 \\
                                     & F1 & 88.92 & 88.48 & 88.66 & 88.76 & 88.99 & 88.76 & 0.20   \\ \bottomrule
\end{tabular}
}
\caption{Results of multiple runs under same training and hyper-parameter setting with different random seeds.}
\label{table:multiple_runs}
\end{table}

\subsection*{Joint Training Results}

We first evaluate \ours{} joint training effect to the QA task by comparing \model{BERT}{base} models and their variants with \ours{} augmented.
In their \ours{} versions, the \ours{} modules only participate in the training process and are removed in the inference task.
\Tbl{table:speedup_datasets} shows the result on multiple QA datasets.
\ours{} outperforms the baseline training objective on all datasets evaluated and exceeds with $0.58\%$ F1 score on average.
This suggests that the \ours{} objective is consistent with the QA objective and even improves its accuracy.
The results show the wide applicability of our method to different datasets with varying difficulty and complexity.

We further show the robustness when using the \ours{} joint training as an add-on module.
\Tbl{table:multiple_runs} shows the result of multiple runs using the identical optimization setting with different random seeds.
By introducing the \ours{} loss in Training, the QA accuracy of the backbone model is improved for $0.4$ on exact match and $0.32$ on F1 score.
And triggering \ours{} always surpasses the backbone model with the same setting.
This is because the extra training objective provides direct gradient signals to the attention mechanism and regularizes its value distribution.

\begin{figure}[t]
  \centering
    \includegraphics[width=1\linewidth]{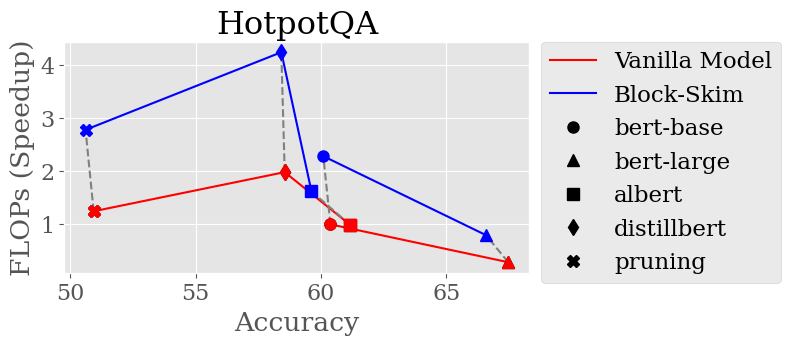}
    \includegraphics[width=1\linewidth]{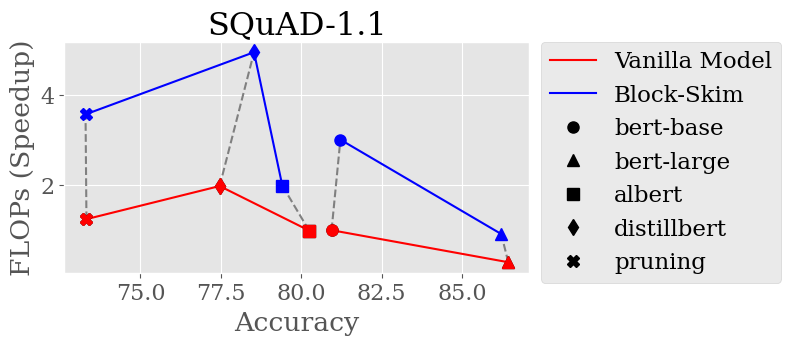}
  \caption{FLOPs speedup of different models and model compression methods with \ours{} on SQuAD and HotpotQA datasets. The FLOPs are normalized to \model{BERT}{base} result of $48.32G$ FLOPs result. The results of vanilla models with different size, model compression algorithms and \ours{} augmented methods are grouped together.}
  \label{fig:speedup_compression}
  \vspace*{-0.3cm}
\end{figure}

% \begin{figure}[t]
%   \centering
%   \includegraphics[width=\linewidth]{fig/longtail_full.pdf}
%   \caption{Breakdown of remained blocks numbers referring to input sequence length. As expected, more blocks are skimmed when processing longer sequence.}
%   \label{fig:longtail}
% \end{figure}

% \begin{figure}[h]
%   \centering
%   \includegraphics[width=0.5\linewidth]{fig/acc_bert_base_layer2.png}
%   \includegraphics[width=0.5\linewidth]{fig/acc_bert_base_layer4.png}
% \end{figure}

\subsection*{Inference Speedup Results} \label{sec:speedup}

\paragraph{Results on Various Datasets.}
The FLOPs speedup result normalized to \model{BERT}{base} model is demonstrated in \Tbl{table:speedup_datasets}.
\ours{} achieves $2.59\times$ speedup on average with a minor accuracy degradation of $0.23$ on different datasets evaluated.
On the multi-hop QA dataset HotpotQA, our method also achieves $2.28$ times speedup.
The results show that the proposed \ours{} method is capable to identify the semantic redundancy with attention information.

\paragraph{Comparison to Vanilla BERT Baseline.}
\ours{} improves the \model{BERT}{base} model inference latency by $3.1\times$ and $2.4\times$ respectively on SQuAD and HotpotQA datasets.
When treating the model size settings of vanilla BERT model as a trade-off between accuracy and complexity, \ours{} improves this trade-off by a margin.
As shown in \Fig{fig:speedup_compression} our method accelerate \model{BERT}{large} as fast as the vanilla \model{BERT}{base} model but with a much higher accuracy.
In specific, the latency of vanilla \model{BERT}{large} model is $3.47\times$ of \model{BERT}{base}, and our method reduces the gap to $1.09\times$ on SQuAD dataset, which translates to the $3.18\times$ speedup.

\paragraph{Compatibility to Model Compression Methods.}
We compare the \ours{}'s compatibility to other model compression methods with \Fig{fig:speedup_compression}.
These model compression methods trade accuracy for computation complexity to different extents.
For example, distilling 12-layer \model{BERT}{base} model to 6 layers results in a $2\%$ accuracy decrease and 2 times speedup.
With \ours{} method appended to this model, the methods can be further accelerated with no or minor accuracy loss.
Specifically, using \ours{} with DistilBERT achieves $5\times$ speedup compared to the vanilla BERT model.
And even with head-pruning reducing the attention information, \ours{} is also compatible and achieves over $2\times$ speedup.
Even though still compatible, \ours{} gets less acceleration on ALBERT models.
We suggest that sharing parameters of attention mechanism makes it harder to optimize with extra \ours{} objective.
As the proposed \ours{} method aims to reduce the input sequence dimension semantic redundancy, it is compatible to these model compression methods focusing on model redundancy theoretically.
By designing \ours{} not to modify the backbone model, our method is generally applicable to these algorithms as well as other model pruning methods~\cite{guo2020accelerating, path_cvpr19, ptolemy_micro20}.

\ours{} achieves close speedup with less accuracy degradation compared to Deformer and more speedup with similar accuracy degradation compared to Length-Adaptive Transformer on SQuAD-1.1 dataset.
This suggests \ours{} captures the runtime semantic redundancy better.
Although this also makes it only applicable to QA tasks.
% \Fig{fig:longtail} shows the ratio of tokens that remain when skimming at different layers.

% The speedup of proposed input rejection method at layer 4 and $0.5$ rejection threshold is described in the following table.
% Generally, we obtained 1.4$\times$ speedup on CPU.
% When processed on GPU with batch size 1, the GPU computation resources are not fully utilized.
% As a result, the model can not be accelerated even if input dimension are reduced.
% To calibrate, we restrict the GPU resource at 50$\%$ and the method achieves $1.62\times$ acceleration.

\begin{table}[]
    \Huge
    \centering
    \resizebox{\linewidth}{!}{
    \begin{tabular}{clcccc|cc}
    \toprule
    \multirow{2}{*}{ID} & \multirow{2}{*}{Description}                          & \multirowcell{2}{Update\\ Transformer} & \multirowcell{2}{Skim\\ Training} & \multirowcell{2}{\ours{}\\ Module}   & \multirowcell{2}{Block\\ Size} & \multicolumn{2}{c}{QA}  \\ \cmidrule(lr){7-8}  
                        &                                            &                                    &                                                    &                            &                             & EM         & F1             \\ \cmidrule(lr){1-8}
\multicolumn{8}{c}{SQuAD} \\  \cmidrule(lr){1-8} 
    1                   & \multicolumn{1}{l|}{Baseline}                &\checkmark              & -                              & -                                           & -                           & 80.92      & 88.32                  \\
    2                   & \multicolumn{1}{l|}{\ours{}}              &\checkmark              &                                &\checkmark                       & 32                          & 81.52      & 88.92           \\
    3                   & \multicolumn{1}{l|}{Freeze Transformer}                &                                    &                                &\checkmark                        & 32                          & 80.92      & 88.32               \\  
    4                   & \multicolumn{1}{l|}{Skim Traning} &\checkmark              &\checkmark          &\checkmark                         & 32                          & 79.27      & 86.83                   \\ \cmidrule(lr){1-8}
    5                   & \multicolumn{1}{l|}{Block Size 1}          &\checkmark              &                                &\checkmark                         & 1                           & 81.22      & 88.60               \\
    6                   & \multicolumn{1}{l|}{Block Size 8}          &\checkmark              &                                &\checkmark                        & 8                           & 81.25      & 88.63             \\
    7                   & \multicolumn{1}{l|}{Block Size 16}         &\checkmark              &                                &\checkmark                       & 16                          & 81.35      & 88.75               \\
    8                  & \multicolumn{1}{l|}{Block Size 64}         &\checkmark              &                                &\checkmark                        & 64                          & 81.39      & 88.65             \\
    9                  & \multicolumn{1}{l|}{Block Size 128}        &\checkmark              &                                &\checkmark                       & 128                         & 80.90      & 88.33              \\ \cmidrule(lr){1-8} 
    \multicolumn{8}{c}{HotpotQA} \\ \cmidrule(lr){1-8}
    10                   & \multicolumn{1}{l|}{Baseline}                &\checkmark              & -                              & -                                           & -                           & 60.37      & 74.39                 \\
    11                   & \multicolumn{1}{l|}{\ours{}}              &\checkmark              &                                &\checkmark                       & 32                          & 60.54      & 74.88          \\
    12                   & \multicolumn{1}{l|}{Evidence Loss}                &            \checkmark                      &                                &\checkmark                        & 32                          & 60.78     & 74.85               \\      
    
    \bottomrule
\end{tabular}}
\caption{Ablation studies of the \ours{} components with \model{BERT}{base} backbone model on SQuAD and HotpotQA datasets.}
\label{table:ablation}
\vspace*{-0.2cm}
\end{table}

\subsection*{Ablation Study} \label{sec:ablation}

We design a series ablation experiment of components in \ours{} to study their individual effect.
The experiments are performed based on the same setting.
We report the detailed results in \Tbl{table:ablation}, and summarize the key findings as follows.

 \paragraph{ID-3.} Instead of joint training, we perform a two-step training. We first perform the fine-tuning for the QA task. We then perform the \ours{} training with the baseline QA model frozen. In other words, we only use the \ours{} objective and only update the weights in the \ours{} modules. Therefore, the QA accuracy remains the same as the baseline model, which is lower than the joint training (ID-3). Meanwhile, the \ours{} classifier also has a lower accuracy than the joint training especially at layer 6.
 
  \paragraph{ID-4} We skim blocks during the joint \ours{} QA training process. Because the mis-skimmed blocks may confuse the QA optimization, it leads to a considerable accuracy loss.

  \paragraph{ID-5-ID-9.} We study the impact of different block sizes. Specifically, when the block size is 1, it is equivalent to skim at the token granularity. Our experimental result shows that the accuracy of \ours{} classifier is better when the block size is larger. On the other hand, a larger block size also leads to less number of blocks and therefore the performance speedup becomes limited. To this end, we choose the block size of 32 as a design sweet spot.

  \paragraph{ID-11-ID-12.} We evaluate the applicability of \ours{} to multi-hop QA task with HotpotQA dataset. As introduced in , we add supporting facts (i.e., evidence) for each question to the \ours{} objective in the ID-12 experiment by labelling evidence blocks to 1 in Eq.2 for the skim predictor modules.
  This leads to a higher QA accuracy.
  But the average accuracy of skim predictors at all layers is worse, which is $86.08\%$ compared to $92.67\%$.
  This ablation experiment shows that our single-task multi-objective joint training is already able to capture the evidence information, rendering explicitly adding it to the training unnecessary.

%% file: sec/conclusion.tex
\section*{Conclusion}

In this work, we propose a plug-and-play \ours{} module to Transformer and its variants for efficient QA processing.
We empirically demonstrate that the attention mechanism can provide instructive information for locating the answer span. 
%In fact, we find that an attention map can distinguish between answers and other tokens.
Leveraging this insight, we propose to learn the attention in a supervised manner, which terminates irrelevant blocks at early layers, significantly reducing the computations.
Besides, the proposed \ours{} training objective provides attention mechanism with extra learning signal and improves QA accuracy on all datasets and models we evaluated. With the use of \ours{} module, such distinction is strengthened in a supervised fashion.
This idea may be also applicable to other tasks and architectures.

%{
%\vspace{.2cm}
%\small

%}

%% file: sec/appendix/attn_distrib.tex
\section*{Appendix Attention Distribution} \label{sec:distrib}
We show the full results of attention weight value distribution discussed in Fig.2.
\Fig{fig:full_avg_attn} shows that deeper layers have more distinguishable patterns.
% However, pruning input sequence at deeper positions provides less acceleration gain.

%% file: sec/acknowledgement.tex
% \begin{figure}[H] 
%   \centering

%   \includegraphics[width=\linewidth]{fig//prof_attn_distrib_3.png}
%   \includegraphics[width=\linewidth]{fig//prof_attn_distrib_5.png}
%   \includegraphics[width=\linewidth]{fig//prof_attn_distrib_6.png}
%   \includegraphics[width=\linewidth]{fig//prof_attn_distrib_7.png}
%   \includegraphics[width=\linewidth]{fig//prof_attn_distrib_8.png}

%   \includegraphics[width=\linewidth]{fig//prof_attn_distrib_10.png}
%   \includegraphics[width=\linewidth]{fig//prof_attn_distrib_11.png}
%   \caption{Attention weight value distribution comparison on the answer and irrelevant tokens. The attention heatmaps are profiled on the development set of SQuAD dataset with a \model{BERT}{base} model with 12 layers and 12 attention heads per layer.}
%   \label{fig:full_avg_attn_continued}
% \end{figure}

\section*{Acknowledgements}
This work was supported by the National Key R\&D Program of China under Grant 2019YFF0302600, and the National Natural Science Foundation of China (NSFC) grant (62072297, 62106143, and 61832006).
We would like to thank the anonymous reviewers for their thoughtful comments and constructive suggestions. Zhouhan Lin is also supported by Shanghai Pujiang Program.
We also thank Yuxian Qiu and Kexin Li with whom we have inspiring discussions on the evaluation experiment design.
Finally, we thank Zhihui Zhang for helping the presentation and visualization of experimental results.
Jingwen Leng and Zhouhan Lin are the corresponding authors of this paper.

\begin{figure}[H] 
  \centering
  \includegraphics[width=0.8\linewidth]{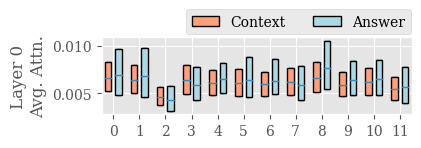}
  \includegraphics[width=0.8\linewidth]{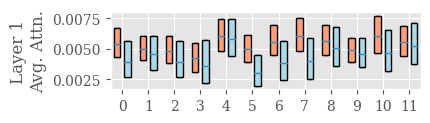}
  \includegraphics[width=0.8\linewidth]{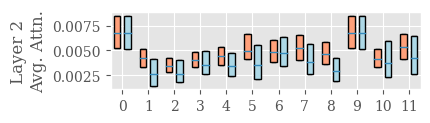}
  \includegraphics[width=0.8\linewidth]{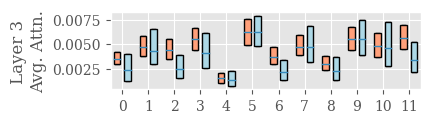}
  \includegraphics[width=0.8\linewidth]{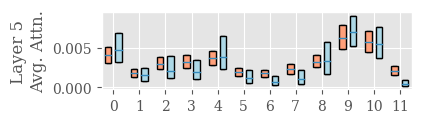}
  \includegraphics[width=0.8\linewidth]{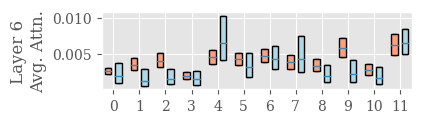}
  \includegraphics[width=0.8\linewidth]{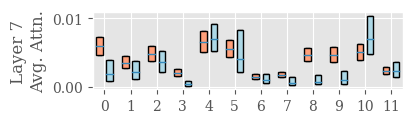}
   \includegraphics[width=0.8\linewidth]{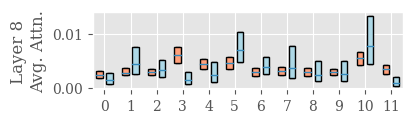}
  \includegraphics[width=0.8\linewidth]{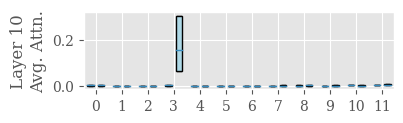}
  \includegraphics[width=0.8\linewidth]{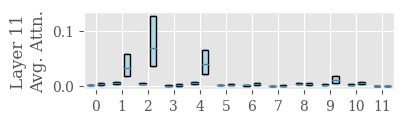}
  \caption{Attention weight value distribution comparison on the answer and irrelevant tokens. The attention heatmaps are profiled on the development set of SQuAD dataset with a \model{BERT}{base} model with 12 layers and 12 attention heads per layer.}
  \label{fig:full_avg_attn}
\end{figure}

%% file: manuscript.bbl
\begin{thebibliography}{43}
\providecommand{\natexlab}[1]{#1}

\bibitem[{Beltagy, Peters, and Cohan(2020)}]{beltagy2020longformer}
Beltagy, I.; Peters, M.~E.; and Cohan, A. 2020.
\newblock Longformer: The long-document transformer.
\newblock \emph{arXiv preprint arXiv:2004.05150}.

\bibitem[{Campos et~al.(2017)Campos, Jou, Gir{\'{o}}{-}i{-}Nieto, Torres, and
  Chang}]{DBLP:journals/corr/abs-1708-06834}
Campos, V.; Jou, B.; Gir{\'{o}}{-}i{-}Nieto, X.; Torres, J.; and Chang, S.
  2017.
\newblock Skip {RNN:} Learning to Skip State Updates in Recurrent Neural
  Networks.
\newblock \emph{CoRR}, abs/1708.06834.

\bibitem[{Cao et~al.(2020)Cao, Trivedi, Balasubramanian, and
  Balasubramanian}]{cao2020deformer}
Cao, Q.; Trivedi, H.; Balasubramanian, A.; and Balasubramanian, N. 2020.
\newblock Deformer: Decomposing pre-trained transformers for faster question
  answering.
\newblock \emph{arXiv preprint arXiv:2005.00697}.

\bibitem[{Clark et~al.(2019)Clark, Khandelwal, Levy, and
  Manning}]{clark2019does}
Clark, K.; Khandelwal, U.; Levy, O.; and Manning, C.~D. 2019.
\newblock What Does BERT Look at? An Analysis of BERT’s Attention.
\newblock In \emph{Proceedings of the 2019 ACL Workshop BlackboxNLP: Analyzing
  and Interpreting Neural Networks for NLP}, 276--286.

\bibitem[{Dai et~al.(2020)Dai, Lai, Yang, and Le}]{dai2020funnel}
Dai, Z.; Lai, G.; Yang, Y.; and Le, Q.~V. 2020.
\newblock Funnel-Transformer: Filtering out Sequential Redundancy for Efficient
  Language Processing.
\newblock \emph{arXiv preprint arXiv:2006.03236}.

\bibitem[{Dehghani et~al.(2018)Dehghani, Gouws, Vinyals, Uszkoreit, and
  Kaiser}]{dehghani2018universal}
Dehghani, M.; Gouws, S.; Vinyals, O.; Uszkoreit, J.; and Kaiser, L. 2018.
\newblock Universal Transformers.
\newblock In \emph{International Conference on Learning Representations}.

\bibitem[{Devlin et~al.(2018)Devlin, Chang, Lee, and
  Toutanova}]{devlin2018bert}
Devlin, J.; Chang, M.-W.; Lee, K.; and Toutanova, K. 2018.
\newblock Bert: Pre-training of deep bidirectional transformers for language
  understanding.
\newblock \emph{arXiv preprint arXiv:1810.04805}.

\bibitem[{Dunn et~al.(2017)Dunn, Sagun, Higgins, Guney, Cirik, and
  Cho}]{dunn2017searchqa}
Dunn, M.; Sagun, L.; Higgins, M.; Guney, V.~U.; Cirik, V.; and Cho, K. 2017.
\newblock Searchqa: A new q\&a dataset augmented with context from a search
  engine.
\newblock \emph{arXiv preprint arXiv:1704.05179}.

\bibitem[{{Gan} et~al.(2020){Gan}, {Qiu}, {Leng}, {Guo}, and
  {Zhu}}]{ptolemy_micro20}
{Gan}, Y.; {Qiu}, Y.; {Leng}, J.; {Guo}, M.; and {Zhu}, Y. 2020.
\newblock {Ptolemy: Architecture Support for Robust Deep Learning}.
\newblock In \emph{2020 53rd Annual IEEE/ACM International Symposium on
  Microarchitecture (MICRO)}.

\bibitem[{Goyal et~al.(2020)Goyal, Choudhary, Chakaravarthy, ManishRaje,
  Sabharwal, and Verma}]{goyal2020power}
Goyal, S.; Choudhary, A.~R.; Chakaravarthy, V.; ManishRaje, S.; Sabharwal, Y.;
  and Verma, A. 2020.
\newblock PoWER-BERT: Accelerating BERT inference for Classification Tasks.
\newblock \emph{arXiv preprint arXiv:2001.08950}.

\bibitem[{Guan et~al.(2020)Guan, Leng, Li, Chen, and Guo}]{guan2020far}
Guan, Y.; Leng, J.; Li, C.; Chen, Q.; and Guo, M. 2020.
\newblock How Far Does BERT Look At: Distance-based Clustering and Analysis of
  BERT $'$ s Attention.
\newblock \emph{arXiv preprint arXiv:2011.00943}.

\bibitem[{Guo et~al.(2020)Guo, Hsueh, Leng, Qiu, Guan, Wang, Jia, Li, Guo, and
  Zhu}]{guo2020accelerating}
Guo, C.; Hsueh, B.~Y.; Leng, J.; Qiu, Y.; Guan, Y.; Wang, Z.; Jia, X.; Li, X.;
  Guo, M.; and Zhu, Y. 2020.
\newblock Accelerating Sparse DNN Models without Hardware-Support via Tile-Wise
  Sparsity.
\newblock \emph{arXiv preprint arXiv:2008.13006}.

\bibitem[{Guo et~al.(2022)Guo, Qiu, Leng, Gao, Zhang, Liu, Yang, Zhu, and
  Guo}]{guo2022squant}
Guo, C.; Qiu, Y.; Leng, J.; Gao, X.; Zhang, C.; Liu, Y.; Yang, F.; Zhu, Y.; and
  Guo, M. 2022.
\newblock {SQ}uant: On-the-Fly Data-Free Quantization via Diagonal Hessian
  Approximation.
\newblock In \emph{International Conference on Learning Representations}.

\bibitem[{Hahnloser and Seung(2001)}]{hahnloser2001permitted}
Hahnloser, R.~H.; and Seung, H.~S. 2001.
\newblock Permitted and forbidden sets in symmetric threshold-linear networks.
\newblock In \emph{Advances in neural information processing systems},
  217--223.

\bibitem[{Hansen et~al.(2018)Hansen, Hansen, Alstrup, Simonsen, and
  Lioma}]{hansen2018neural}
Hansen, C.; Hansen, C.; Alstrup, S.; Simonsen, J.~G.; and Lioma, C. 2018.
\newblock Neural Speed Reading with Structural-Jump-LSTM.
\newblock In \emph{International Conference on Learning Representations}.

\bibitem[{Hochreiter and Schmidhuber(1997)}]{hochreiter1997long}
Hochreiter, S.; and Schmidhuber, J. 1997.
\newblock Long short-term memory.
\newblock \emph{Neural computation}, 9(8): 1735--1780.

\bibitem[{Huang et~al.(2020)Huang, Xu, Hu, Wang, Qiu, Fu, Zhao, Peng, and
  Wang}]{huang2020recent}
Huang, Z.; Xu, S.; Hu, M.; Wang, X.; Qiu, J.; Fu, Y.; Zhao, Y.; Peng, Y.; and
  Wang, C. 2020.
\newblock Recent Trends in Deep Learning Based Open-Domain Textual Question
  Answering Systems.
\newblock \emph{IEEE Access}, 8: 94341--94356.

\bibitem[{Ioffe and Szegedy(2015)}]{ioffe2015batch}
Ioffe, S.; and Szegedy, C. 2015.
\newblock Batch Normalization: Accelerating Deep Network Training by Reducing
  Internal Covariate Shift.
\newblock In \emph{International Conference on Machine Learning}, 448--456.

\bibitem[{Joshi et~al.(2017)Joshi, Choi, Weld, and
  Zettlemoyer}]{joshi2017triviaqa}
Joshi, M.; Choi, E.; Weld, D.~S.; and Zettlemoyer, L. 2017.
\newblock TriviaQA: A Large Scale Distantly Supervised Challenge Dataset for
  Reading Comprehension.
\newblock In \emph{Proceedings of the 55th Annual Meeting of the Association
  for Computational Linguistics (Volume 1: Long Papers)}, 1601--1611.

\bibitem[{Kim and Cho(2020)}]{kim2020length}
Kim, G.; and Cho, K. 2020.
\newblock Length-Adaptive Transformer: Train Once with Length Drop, Use Anytime
  with Search.
\newblock \emph{arXiv preprint arXiv:2010.07003}.

\bibitem[{Kitaev, Kaiser, and Levskaya(2019)}]{kitaev2019reformer}
Kitaev, N.; Kaiser, L.; and Levskaya, A. 2019.
\newblock Reformer: The Efficient Transformer.
\newblock In \emph{International Conference on Learning Representations}.

\bibitem[{Kovaleva et~al.(2019)Kovaleva, Romanov, Rogers, and
  Rumshisky}]{kovaleva2019revealing}
Kovaleva, O.; Romanov, A.; Rogers, A.; and Rumshisky, A. 2019.
\newblock Revealing the dark secrets of BERT.
\newblock \emph{arXiv preprint arXiv:1908.08593}.

\bibitem[{Kwiatkowski et~al.(2019)Kwiatkowski, Palomaki, Redfield, Collins,
  Parikh, Alberti, Epstein, Polosukhin, Devlin, Lee
  et~al.}]{kwiatkowski2019natural}
Kwiatkowski, T.; Palomaki, J.; Redfield, O.; Collins, M.; Parikh, A.; Alberti,
  C.; Epstein, D.; Polosukhin, I.; Devlin, J.; Lee, K.; et~al. 2019.
\newblock Natural questions: a benchmark for question answering research.
\newblock \emph{Transactions of the Association for Computational Linguistics},
  7: 453--466.

\bibitem[{Lan et~al.(2019)Lan, Chen, Goodman, Gimpel, Sharma, and
  Soricut}]{lan2019albert}
Lan, Z.; Chen, M.; Goodman, S.; Gimpel, K.; Sharma, P.; and Soricut, R. 2019.
\newblock ALBERT: A Lite BERT for Self-supervised Learning of Language
  Representations.
\newblock In \emph{International Conference on Learning Representations}.

\bibitem[{Lin et~al.(2017)Lin, Feng, Santos, Yu, Xiang, Zhou, and
  Bengio}]{lin2017structured}
Lin, Z.; Feng, M.; Santos, C. N.~d.; Yu, M.; Xiang, B.; Zhou, B.; and Bengio,
  Y. 2017.
\newblock A structured self-attentive sentence embedding.
\newblock \emph{arXiv preprint arXiv:1703.03130}.

\bibitem[{Liu et~al.(2019)Liu, Ott, Goyal, Du, Joshi, Chen, Levy, Lewis,
  Zettlemoyer, and Stoyanov}]{liu2019roberta}
Liu, Y.; Ott, M.; Goyal, N.; Du, J.; Joshi, M.; Chen, D.; Levy, O.; Lewis, M.;
  Zettlemoyer, L.; and Stoyanov, V. 2019.
\newblock RoBERTa: A Robustly Optimized BERT Pretraining Approach.

\bibitem[{Liu(2020)}]{torchprofile}
Liu, Z. 2020.
\newblock Torchprofile.
\newblock \url{https://github.com/zhijian-liu/torchprofile/}.

\bibitem[{Michel, Levy, and Neubig(2019)}]{michel2019sixteen}
Michel, P.; Levy, O.; and Neubig, G. 2019.
\newblock Are Sixteen Heads Really Better than One?
\newblock \emph{Advances in Neural Information Processing Systems}, 32:
  14014--14024.

\bibitem[{Qiu et~al.(2019)Qiu, Leng, Guo, Chen, Li, Guo, and Zhu}]{path_cvpr19}
Qiu, Y.; Leng, J.; Guo, C.; Chen, Q.; Li, C.; Guo, M.; and Zhu, Y. 2019.
\newblock {Adversarial Defense Through Network Profiling Based Path
  Extraction}.
\newblock In \emph{Proceedings of the IEEE/CVF Conference on Computer Vision
  and Pattern Recognition (CVPR)}.

\bibitem[{Rajpurkar et~al.(2016)Rajpurkar, Zhang, Lopyrev, and
  Liang}]{rajpurkar2016squad}
Rajpurkar, P.; Zhang, J.; Lopyrev, K.; and Liang, P. 2016.
\newblock SQuAD: 100, 000+ Questions for Machine Comprehension of Text.
\newblock In \emph{EMNLP}.

\bibitem[{Rumelhart, Hinton, and Williams(1986)}]{rumelhart1986learning}
Rumelhart, D.~E.; Hinton, G.~E.; and Williams, R.~J. 1986.
\newblock Learning representations by back-propagating errors.
\newblock \emph{nature}, 323(6088): 533--536.

\bibitem[{Sanh et~al.(2019)Sanh, Debut, Chaumond, and
  Wolf}]{sanh2019distilbert}
Sanh, V.; Debut, L.; Chaumond, J.; and Wolf, T. 2019.
\newblock DistilBERT, a distilled version of BERT: smaller, faster, cheaper and
  lighter.
\newblock \emph{arXiv preprint arXiv:1910.01108}.

\bibitem[{Seo et~al.(2018)Seo, Min, Farhadi, and Hajishirzi}]{seo2018neural}
Seo, M.; Min, S.; Farhadi, A.; and Hajishirzi, H. 2018.
\newblock Neural Speed Reading via Skim-RNN.
\newblock In \emph{International Conference on Learning Representations}.

\bibitem[{Tay et~al.(2020)Tay, Dehghani, Bahri, and Metzler}]{tay2020efficient}
Tay, Y.; Dehghani, M.; Bahri, D.; and Metzler, D. 2020.
\newblock Efficient Transformers: A Survey.
\newblock \emph{arXiv e-prints}, arXiv--2009.

\bibitem[{Trischler et~al.(2016)Trischler, Wang, Yuan, Harris, Sordoni,
  Bachman, and Suleman}]{trischler2016newsqa}
Trischler, A.; Wang, T.; Yuan, X.; Harris, J.; Sordoni, A.; Bachman, P.; and
  Suleman, K. 2016.
\newblock NEWSQA: A MACHINE COMPREHENSION DATASET.

\bibitem[{Vaswani et~al.(2017)Vaswani, Shazeer, Parmar, Uszkoreit, Jones,
  Gomez, Kaiser, and Polosukhin}]{vaswani2017attention}
Vaswani, A.; Shazeer, N.; Parmar, N.; Uszkoreit, J.; Jones, L.; Gomez, A.~N.;
  Kaiser, {\L}.; and Polosukhin, I. 2017.
\newblock Attention is all you need.
\newblock In \emph{Advances in neural information processing systems},
  5998--6008.

\bibitem[{Wang and Zhang(2020)}]{wang2020q}
Wang, C.; and Zhang, X. 2020.
\newblock Q-BERT: A BERT-based Framework for Computing SPARQL Similarity in
  Natural Language.
\newblock In \emph{Companion Proceedings of the Web Conference 2020}, 65--66.

\bibitem[{Wolf et~al.(2019)Wolf, Debut, Sanh, Chaumond, Delangue, Moi, Cistac,
  Rault, Louf, Funtowicz, Davison, Shleifer, von Platen, Ma, Jernite, Plu, Xu,
  Scao, Gugger, Drame, Lhoest, and Rush}]{Wolf2019HuggingFacesTS}
Wolf, T.; Debut, L.; Sanh, V.; Chaumond, J.; Delangue, C.; Moi, A.; Cistac, P.;
  Rault, T.; Louf, R.; Funtowicz, M.; Davison, J.; Shleifer, S.; von Platen,
  P.; Ma, C.; Jernite, Y.; Plu, J.; Xu, C.; Scao, T.~L.; Gugger, S.; Drame, M.;
  Lhoest, Q.; and Rush, A.~M. 2019.
\newblock HuggingFace's Transformers: State-of-the-art Natural Language
  Processing.
\newblock \emph{ArXiv}, abs/1910.03771.

\bibitem[{Wu et~al.(2019)Wu, Liu, Lin, Lin, and Han}]{wu2019lite}
Wu, Z.; Liu, Z.; Lin, J.; Lin, Y.; and Han, S. 2019.
\newblock Lite Transformer with Long-Short Range Attention.
\newblock In \emph{International Conference on Learning Representations}.

\bibitem[{Yang et~al.(2018)Yang, Qi, Zhang, Bengio, Cohen, Salakhutdinov, and
  Manning}]{yang2018hotpotqa}
Yang, Z.; Qi, P.; Zhang, S.; Bengio, Y.; Cohen, W.~W.; Salakhutdinov, R.; and
  Manning, C.~D. 2018.
\newblock HotpotQA: A Dataset for Diverse, Explainable Multi-hop Question
  Answering.
\newblock In \emph{EMNLP}.

\bibitem[{Yu, Lee, and Le(2017)}]{yu2017learning}
Yu, A.~W.; Lee, H.; and Le, Q. 2017.
\newblock Learning to Skim Text.
\newblock In \emph{Proceedings of the 55th Annual Meeting of the Association
  for Computational Linguistics (Volume 1: Long Papers)}.

\bibitem[{Zaheer et~al.(2020)Zaheer, Guruganesh, Dubey, Ainslie, Alberti,
  Ontanon, Pham, Ravula, Wang, Yang et~al.}]{zaheer2020big}
Zaheer, M.; Guruganesh, G.; Dubey, A.; Ainslie, J.; Alberti, C.; Ontanon, S.;
  Pham, P.; Ravula, A.; Wang, Q.; Yang, L.; et~al. 2020.
\newblock Big bird: Transformers for longer sequences.
\newblock \emph{arXiv preprint arXiv:2007.14062}.

\bibitem[{Zhou et~al.(2020)Zhou, Xu, Ge, McAuley, Xu, and Wei}]{zhou2020bert}
Zhou, W.; Xu, C.; Ge, T.; McAuley, J.; Xu, K.; and Wei, F. 2020.
\newblock BERT Loses Patience: Fast and Robust Inference with Early Exit.
\newblock \emph{arXiv}, arXiv--2006.

\end{thebibliography}
